\documentclass[journal]{IEEEtran}
\usepackage{amssymb,amsmath,bm}
\usepackage[lofdepth,lotdepth]{subfig}

\usepackage{framed,multirow}

\usepackage{amssymb}
\usepackage{latexsym}

\usepackage{url}
\usepackage{xcolor}
\usepackage{times}
\usepackage{epsfig}
\usepackage{graphicx}
\usepackage{amsmath}
\usepackage{amssymb}
\usepackage{algorithm,algorithmic}
\usepackage{url}
\usepackage{array}
\usepackage{bm}
\usepackage{verbatim}
\usepackage{placeins}
\usepackage{multicol}
\usepackage{lipsum}
\usepackage{tikz}
\usepackage{caption}
\usepackage{subfig}

\usepackage{multirow}

\graphicspath{ {images/} }

\ifCLASSINFOpdf
\else
\fi

\begin{document}
%
\title{Universal Image Manipulation Detection using Deep Siamese Convolutional Neural Network}
%
%
%

\author{\IEEEauthorblockN{Aniruddha Mazumdar\IEEEauthorrefmark{1}, Jaya Singh, Yosha Singh Tomar, and 
        Prabin Kumar Bora}\\ 
\IEEEauthorblockA{Department of Electronics and Electrical Engineering\\
	Indian Institute of Technology Guwahati, Assam, India - 781039\\
	Email: \IEEEauthorrefmark{1}m.aniruddha@iitg.ac.in}}

\maketitle

\begin{abstract}
Detection of different types of image editing operations carried out on an image is an important problem in image forensics. It gives the information about the processing history of an image, and also can expose forgeries present in an image. There have been few methods proposed to detect different types of image editing operations in a single framework. However, all the operations have to be known \emph{a priori} in the training phase. But, in real-forensics scenarios it may not be possible to know about the editing operations carried out on an image. To solve this problem, we propose a novel deep learning-based method which can differentiate between different types of image editing operations. The proposed method classifies image patches in a pair-wise fashion as either similarly or differently processed using a deep siamese neural network.
Once the network learns feature that can discriminate between different image editing operations, it can differentiate between different image editing operations not present in the training stage. 
The experimental results show the efficacy of the proposed method in detecting/discriminating different image editing operations.
\end{abstract}


%
\IEEEpeerreviewmaketitle

\section{Introduction}\label{sec:intro}
With the availability of various image editing software, it has become possible to create visually plausible image forgeries with a minimal effort. Because of this, a large number of forged images are now available on the web. These images when used on different platforms, like the electronic and social media, may create tensions in the society. These concerns necessitate the development of image forensics techniques for checking the authenticity of images before using them as critical information.

While creating a forgery, the forged parts are often processed through different image editing operations to make them look visually plausible. For example, in image \emph{splicing} forgery the spliced objects go through different image editing operations, e.g. resizing, rotating, smoothening, contrast enhanecment, compression. Although imperceptible to human eyes, every image editing operation leaves a unique trace of manipulation. These traces are utilized by researchers to detect different types of editing operations performed on images. Different techniques have been proposed to extract features related to the traces left by different editing operations, and which are utilized to check the authenticity of images. For example, in \cite{resampling}, \cite{resampling_2}, \cite{resampling_3} the authors extracted features for detecting the traces left by resizing and resampling operations, in \cite{median_2}, \cite{median_3}, \cite{median_4} features related to median filtering traces are extracted, in \cite{contrast_1}, \cite{contrast_2} features are extracted to detect contrast-enhancement operation, and in \cite{Bianchi_JPEG} JPEG artifact related features are extracted for forgery detection.


Although these methods are good at detecting splicing, methods from each of the categories work only under their respective assumptions about the traces of manipulations left by the forgery process. For example, the median filtering detection methods cannot detect traces left by the resampling operation. To handle this limitation, researchers have focused to develop universal forensics methods, which can detect multiple manipulations in a single framework. The first universal forensics method was proposed by Qiu \emph{et al.} \cite{Qiu_universal}, where different steganalysis features were used to detect different types of image processing operations. The method is based on the observation that different image editing operations destroy the natural statistics of the image pixels present in an authentic image in the same way steganography methods do while manipulating the pixels for embedding a message. Fan \emph{et al.} \cite{fan_WIFS} proposed another general-purpose forensics method for detecting different types of image editing operations. The authors proposed to create a Gaussian mixture model (GMM) of image patches corresponding each editing operation. Then, the average log-likelihood of patches under the different GMMs corresponding to different classes are compared to decide the class of the patches.

Inspired by the success in other computer vision areas, the forensics community has recently focused on applying deep learning-based methods for image manipulation detection. Chen \emph{et al.} \cite{median_CNN} proposed the first deep learning-based median filtering detection method. This is the first deep learning-based image manipulation detection method, where the first layer computes the median filtering residual and the subsequent layers extract and classify the features useful for median filtering detection. Bayar and Stamm \cite{Bayar_CNN} proposed a deep learning-based universal forensics method for detecting different types of image manipulating operations. The image editing features are automatically learned from the training data by employing a convolutional neural network (CNN) \cite{CNN}. The authors proposed a new convolutional layer, which suppresses the image content and enhances features important for detecting different editing operations. 

Although these universal manipulation detection methods perform really well, all the manipulation operations have to be known before training the network. However, there is a large number of image editing operations available in the image editing software, e.g. Adobe Photoshop and GIMP. Also, new image editing operations are being developed and incorporated in these editing software. In addition to that, there may be multiple editing operations performed subsequently to make the spliced parts look similar to the authentic parts. Therefore, it is not practical to incorporate all the editing operations in the training process as required by the existing univeral manipulation detection methods, i.e. \cite{Qiu_universal}, \cite{fan_WIFS}, \cite{Bayar_CNN}. This necessitates the developement of universal forensics method which can not only detect the different image editing operations present in the training stage, but also is capable of generalizing to editing operations not present in the training stage.

This paper proposes a novel deep learning-based forensics method for detecting image manipulations. The proposed method takes two image patches as input, and check whether they come from the same or different manipulation operations. The proposed method is built upon the work of Bayar and Stamm \cite{Bayar_CNN}, where they showed that CNNs are capable of learning accurate image editing features automatically from the training data. However, instead of learning features to classify image patches to different manipulation classes, the proposed method learns the features which can discriminate different image editing operations. The reason for this is that from the forensics point of view it is more informative to check whether two image patches have undergone the same type of manipulations or not than classifying individual image patch. For this, the proposed method employs a deep siamese CNN, which has twin CNNs accepting two image patches as the input and classifies the patch pair as either identically processed (IP) or differently processed (DP).

\section{Background}
In this section, we explain the deep learning-based universal forensics method \cite{Bayar_CNN} proposed by Basar and Stamm as the background. The method is based on the assumption that each image editing operation leaves behind the trace of the particular manipulation. These traces can be detected by examining the relationship between the neighbouring pixels, as any image manipulation destroys the natural statistics of pixels and modifies them in a unique way \cite{Qiu_universal}. To automatically learn the features useful for the detection different manipulation operations, the authors of \cite{Bayar_CNN} proposed to employ a CNN. 

A CNN is a special type of neural networks originally proposed for handwritten digit recognition \cite{CNN}. Since then, they have been successfully used in many other computer vision problems with some variations in its architecture \cite{alexnet}, \cite{vggnet}. A CNN typically contains a stack of multiple layers with nonlinearities which enable it to learn different features from the training data itself. The typical layers present in a CNN are the convolutional layer, the pooling layer, and the fully connected layer. The convolutional layer contains several filters which convolve with the input image in parallel and the element-wise rectified linear unit (ReLU) for non-linear mapping. The output of the convolutional layer is called the feature map, and given by

\begin{equation} 
\begin{aligned}
\textbf{f}_{k}^{\textbf{ }l} = \text{max}(0, \sum_{j}\textbf{f}_{j}^{\textbf{ }l-1}*\textbf{w}_{kj}^{l}+\textbf{b}_{k}^{l})
\end{aligned}
\end{equation}
where, $*$ is the convolution operation, $\textbf{f}_{k}^{\textbf{ }l}$ is the $k$th feature map in layer $l$, $\textbf{w}_{kj}^{l}$ is the filter connecting the $j$th feature map in layer $l-1$ to the 
$k$th feature map in layer $l$ and $\textbf{b}_{k}^{l}$ is the bias for the $k$th feature map in layer $l$. The convolutional layer is followed by a max-pooling layer, which reduces the size of each feature map by taking the maximum value over a region. The stacking of the convolutional and pooling layers one after another enables the CNNs to learn different levels of features at different layers. The initial layers learn low-level features and the final layers learn more problem-specific features. To classify the final features, one or more fully-connected layers are stacked on the top of the final convolutional and pooling layers. The sigmoid non-linearity is used in the fully-connected layer, producing the output in the range of $[0,1]$. The parameters $\textbf{w}$ and $\textbf{b}$ are learnt in the training process using the standard gradient descent-based backpropagation technique \cite{CNN}.

CNNs have proved to be very good in different computer vision tasks, e.g. object detection and recognition \cite{alexnet}. However, they did not perform well when applied directly to image manipulation detection \cite{Chen_Median}. This is because the conventional CNNs capture the image content rather than important forensics features. To suppress the image content and enhance the relationship between neighbouring pixels, Bayar and Stamm proposed a new convolutional layer as the first layer of the CNN. The filters in this new convolutional layer are constrained to learn a set of prediction error filters. The concept of using the prediction error filters in the first layer is motivated from different image forensics and steganalysis methods. Steganalysis methods like the rich models \cite{rich_steg} and the subtractive pixel adjacency matrix (SPAM) \cite{spam_steg} utilise this concept of using different prediction filters for computing the prediction errors, which are later used as features to detect hidden messages present in the stago images. In forensics, Chen \emph{et al.} \cite{Chen_Median} proposed a similar strategy to first extract the median filtering residuals and then using a CNN for the detection of the median filtering operation.

In \cite{Bayar_CNN}, the filters in the first convolutional layer of the CNN are forced to learn a set of prediction error filters by constraining the weights in each of the $K$ filters as
\begin{equation}\label{constrain}
\begin{aligned}
w_{k}^{1}(0,0)=-1 \\
\text{and} \textbf{} \sum_{l,m\neq 0}w_{k}^{1}(l,m)=1
\end{aligned}
\end{equation}
where, $w_{k}^{1}(l,m)$ denotes the weight at position $(l,m)$ of the $k$th filter and $w_{k}^{1}(0,0)$ denotes the weight at the center of the corresponding filter kernel. This procedure is repeated for all the pixels in the image patch by moving the kernels throughout the patch. This prediction error layer extracts the local dependency of pixels with its neighbours, which is the important information from the forensics point of view \cite{Qiu_universal}.

Using this approach, Bayar and Stamm detected four different types image editing operations, namely the Gaussian filtering, the median filtering, the corruption of the image by the AWGN and resampling, with good accuracies.

\section{Proposed Method}

The universal forensics method proposed by Bayar and Stamm \cite{Bayar_CNN} shows that CNNs can automatically learn features important for detecting different image editing operations. However, there are some limitations of the method. For training the CNN, all the image editing operations should be known \emph{a priori}. If an image is edited with an operation other than the ones used for training, the method will not be able to detect it properly. Moreover, in a real forgery, there may be multiple editing operations performed on the image subsequently. In this case also, the method \cite{Bayar_CNN} will fail as the CNN is trained to classify the images only to one of the many editing operations, not combinations of different manipulations. 

To overcome these limitations, we propose a method which checks whether two image patches have undergone the same operation or two different operations. For this, we employ a deep siamese CNN which takes a pair of image patches as input and classify these as either IP or DP. The reasons for the pair-wise classification image patches are as follows:

\begin{enumerate}
	\item The spliced regions in an image may go through different image editing operations than the authentic regions. Therefore, from the forensics point of view, it is more informative to know whether all the patches of an image have been manipulated in the same way or not.
	
	\item The pair-wise classification of patches removes the necessity of classifying the patches into different manipulation classes. This is an important advantage of the proposed method, as it allows the proposed method to classify image patches coming from a different type of manipulation not considered in the training stage.
	
	\item Since, the methods \cite{Qiu_universal}, \cite{fan_WIFS} and \cite{Bayar_CNN} classify the image patches into one of the different but fixed types of manipulations, they are more vulnerable to anti-forensics. This is because the anti-forensics methods can be developed to hide the traces left by each of the operations considered in these methods \cite{median_anti}. On the other hand, the proposed method does not learn any class-specific feature as it is designed to check whether two patches have undergone the same type of manipulation or not. Hence, developing anti-forensics techniques to counter the proposed method will be more difficult.
\end{enumerate}

\begin{figure}[]
	\centering
	\includegraphics[width=9.0cm, height=4cm]{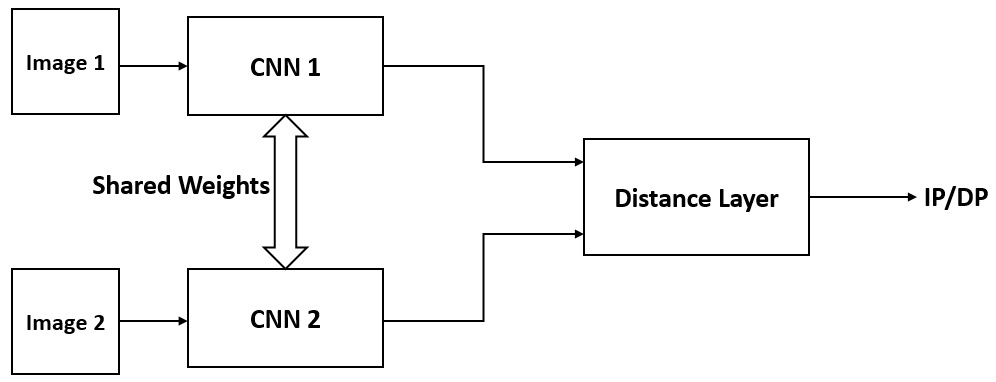}
	
	\caption{Framework of the siamese network to classify patch pairs as IP or DP.} \label{siamese_block}
\end{figure}

Figure \ref{siamese_block} shows the block diagram of the proposed framework. It has twin neural networks CNN1 and CNN2 sharing the same set of weights. It accepts two input images, which are independently processed by CNN1 and CNN2 and then a distance layer \cite{koch} computes a distance metric between the outputs of the twin networks. Because of the sharing of weights, CNN1 and CNN2 map two similar input images to very close points in the feature space. The proposed siamese CNN automatically learns the features that can check whether a pair of images has been similarly or differently manipulated.

\begin{figure*}[]
	\centering
	\includegraphics[width=15.0cm, height=5cm]{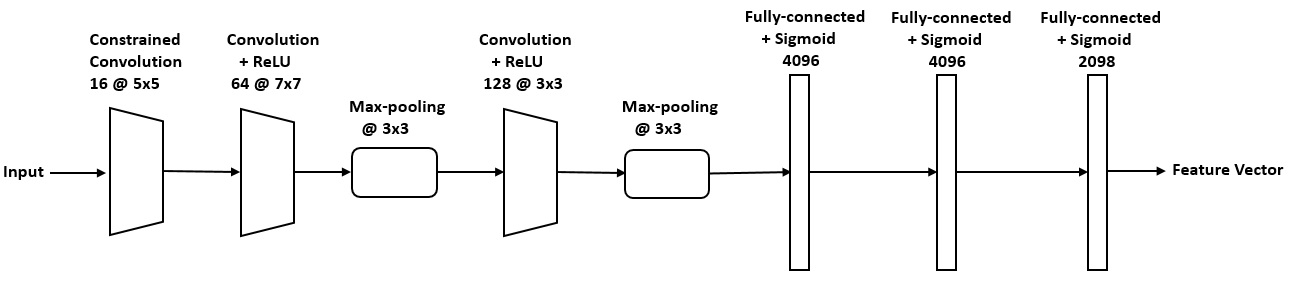} 
	
	\caption{The CNN architecture used in the proposed siamese network.} \label{CNN_block}
\end{figure*} 

\subsection{Network Architecture}
\subsubsection{CNN}
As in \cite{Bayar_CNN}, the first convolutional layer in each of CNN1 and CNN2 is a constrained convolutional layer. The filters in the constrained convolutional layer are forced to learn a set of prediction error filters, which suppress image contents and produce prediction error. Each of CNN1 and CNN2 has the architecture shown in Figure \ref{CNN_block}. It contains $3$ convolutional layers, $2$ max-pooling layers and $3$ fully-connected layers. The block diagram of the CNN is shown in Figure \ref{CNN_block}. The first convolutional layer is the constrained convolutional layer \cite{Bayar_CNN} with $16$ prediction error filters of size $5\times5$ and stride $1$. Its weights follow the constraints given by Equation (\ref{constrain}). This layer is followed by an unconstrained convolutional layer with $64$ filters of size $7\times7$ with stride $2$. The ReLU nonlinearity is applied element-wise to the output of this layer followed by the max-pooling layer with a kernel size $3\times3$ and stride $2$. The output of this layer is fed to another unconstrained convolutional layer with $128$ filters of size $3\times3$ and stride $1$. The ReLU nonlinearity is applied element-wise to the output of this layer. It is followed by a max-pooling layer with a kernel size $3\times3$ and stride $2$. This layer is followed by three fully-connected layers with $4096$, $4096$ and $2048$ neurons respectively. The sigmoid non-linearity is used in each of these layers. The neurons in the fully-connected layers are dropped out \cite{dropout} with a probability of $0.5$ at each iteration of the training process. The output of the final fully-connected layer represents the features learned by the CNN.

\subsubsection{Distance Layer}
Given a pair of image patches $\textbf{x}_{1}$ and $\textbf{x}_{2}$ as input, CNN1 and CNN2 compute the feature vectors $\textbf{f}_{1}$ and $\textbf{f}_{2}$ respectively. A distance layer computes a distance metric between them, which is then fed to a single sigmoidal output neuron. This neuron computes the prediction of the input image patch pair as $p = \sigma(\sum_j \alpha_j \left | f_{1}(j) - f_{2}(j) \right |)$, where $\sigma$ is the sigmoid non-linearity function and $\alpha_{j}$ is a learnable parameter representing the importance of each component of the feature vectors in the classification of the patch-pair.

\subsection{Learning}
The proposed siamese network is a binary classifier with label $y(\textbf{x}_{1}, \textbf{x}_{2})=1$ when both input image patches $\textbf{x}_{1}$ and $\textbf{x}_{2}$ come from the same manipulation class, and $y(\textbf{x}_{1}, \textbf{x}_{2})=0$ when $\textbf{x}_{1}$ and $\textbf{x}_{2}$ come from two different manipulation classes. The network is trained by minimising the average cross-entropy loss function $C$ over a batch of pairs given by \cite{cross_entroy}

\begin{equation} \label{cost_fn}
\begin{aligned}
C = \frac{1}{M}\sum_{i=1}^{M} y(\textbf{x}_1^i, \textbf{x}_2^i) \log p(\textbf{x}_1^i, \textbf{x}_2^i)  \\ + (1-y(\textbf{x}_1^i, \textbf{x}_2^i))\log (1-p(\textbf{x}_1^i, \textbf{x}_2^i))
\end{aligned}
\end{equation}
where $M$ is the number of images in each batch. The parameters of the network are learnt in the training phase by minimising $C$ using the stochastic gradient descent (SGD)-based backpropagation technique. In SGD method, the weights are updated in each iteration using the following equations:
\begin{equation} \label{weight_update}
\begin{aligned}
\textbf{w}_{kj}^{l} = \textbf{w}_{kj}^{l} + \Delta \textbf{w}_{kj}^{l} \\
\Delta \textbf{w}_{kj}^{l} =   \mu \Delta \textbf{w}_{kj}^{l} - \eta \nabla_{\textbf{w}_{kj}^{l}} C - \lambda  \textbf{w}_{kj}^{l}
\end{aligned}
\end{equation}
where, $ \nabla_{\textbf{w}_{kj}^{l}} $ is the gradient of $C$ with respect to the weight matrix $\textbf{w}_{kj}^{l}$, $\eta$ is the learning rate, $\mu$ is the momentum and $\lambda$ is the $L_2$ regularization term.

Once, the network is trained, it is used to detect/discriminate the different image processing operations.

\section{Experiments and Results}
\begin{table*}[!] 
	\centering
	\caption{Different manipulations considered in this paper}
	\label{mainpulation}
	\begin{tabular}{|c|c|}
		\hline
		\textbf{Editing Operation}     & \textbf{Detail}                                                           \\ \hline \hline
		Gaussian blurring & Kernel size = $5\times5$ and standard deviation $(\sigma)$ = $1.1$                   \\ \hline
		Median filtering & Kernel size = $5\times5$                                                \\ \hline
		Resampling       & Scaling factor = $1.5$ and bilinear interpolation                  \\ \hline
		Noise addition   & AWGN with standard deviation $(\sigma)$ = $2$ \\ \hline
		Gamma correction & Parameter $(\gamma)$ = $1.5$                                               \\ \hline
	\end{tabular}
\end{table*}
To train and test the proposed method, a dataset was created using the unprocessed raw images taken from the Dresden Image Database \cite{dresden}. The database contains more than $14,000$ images with resolutions of about $2000\times3000$ captured by $73$ different digital cameras. A set of $1566$ raw images were compressed in the JPEG format with $100\%$ quality factor (QF) and converted to grayscale images by considering only the green channel of the images. We cropped image patches of size $150\times150$ from these images, resulting in $114,000$ unaltered image patches.

The proposed system was implemented using the Python-based Keras \cite{keras} deep learning library on a Tesla K20c GPU with $5$ GB of RAM. The Nadam optimiser \cite{nadam} was used with the parameters set as: $learning \textbf{ } rate (\eta) = 0.002$, $momentum (\mu) = 0.002$ and $decay = 0.005$ and $regularization \textbf{ } term (\lambda) = 0.0001$. We have used the learning rate decay technique to converge to the minimum of $C$ by reducing the fluctuations \cite{decay}. The training batch size was set to $16$ images. We have used the batch normalisation technique \cite{batch_norm} as it helps in achieving faster convergence and higher generalisation accuracy.

\begin{figure*}[] 
	\centering
	\subfloat[]{\includegraphics[width=9.0cm, height=6.5cm]{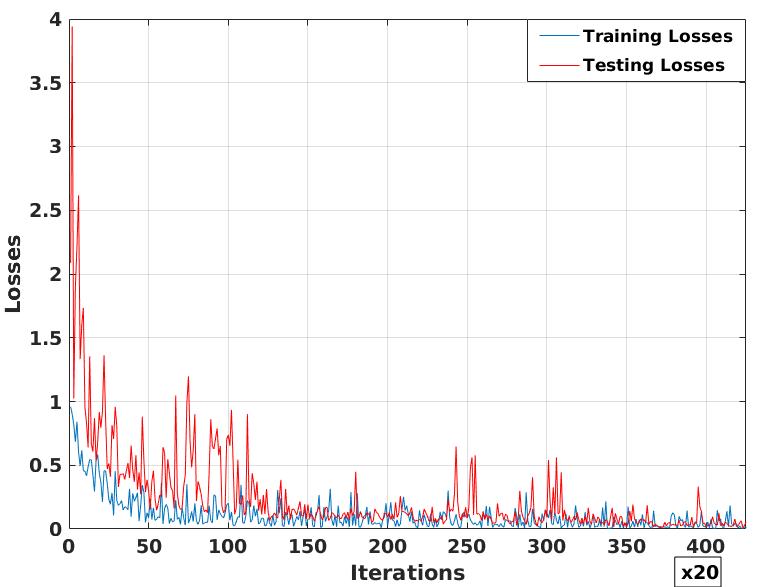}
		\label{loss_iter}} 
	\subfloat[]{\includegraphics[width=9.0cm, height=6.5cm]{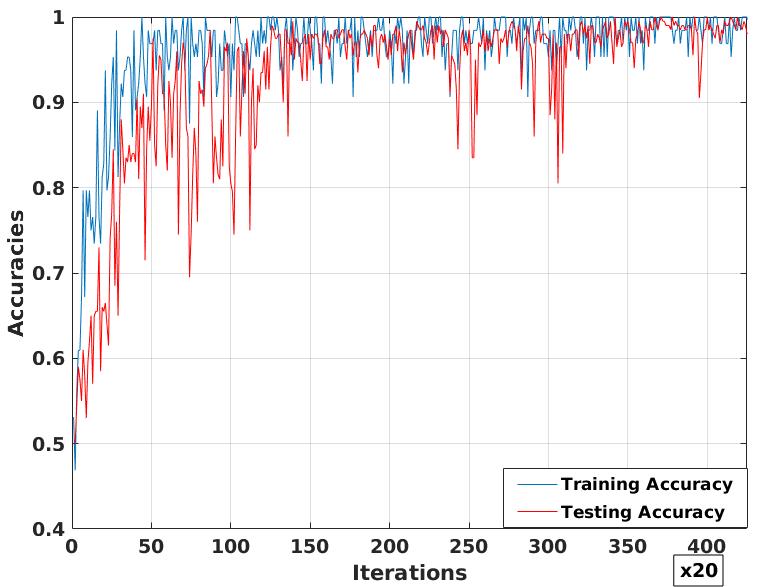}
		\label{acc_iter}}
	\caption{(a) Training and validation loss versus iteration. (b) Training and validation accuracy versus iteration.} \label{plot}
\end{figure*}

\subsection{Manipulation Detection Results}
To test the performance of the proposed siamese network in detecting/discriminating different image editing operations, we have carried out a series of experiments. For this, five different versions of the $114,000$ unaltered patches are created by editing them with the following operations: Gaussian blurring, median filtering, resampling, corrupting with the additive white Gaussian noise (AWGN), gamma correction. The details of the manipulations are listed in Table \ref{mainpulation}. This way, we obtain $114,000$ patches from the altered as well as from each of the editing operations.

In the first experiment, we have trained the network using the image patches coming from four different classes: original, Gaussian blurring, median filtering, and resampling. We randomly selected $40,000$ patches from each class to create the training set. We sample $500,000$ IP pairs of patches randomly where both image patches of a pair come from the same class (i.e. both patches of a pair come either from unaltered class or from the same manipulation class). Similarly, we sample $500,000$ DP pairs randomly, where the two patches of a pair come from two different classes (e.g. one patch may come from Gaussian blurring operation and the other may come from Median filtering operation). To monitor the classification performance of the network during training, we apply it on a validation set which contains $10,000$ IP pairs and $10,000$ DP pairs that are not in the training set. Once, the model is trained we check the performance of the method on a test set which contains $50,000$ IP pairs and $50,000$ DP pairs.

The network was trained for $70,000$ iterations and stopped when it started converging. Figure \ref{loss_iter} shows the training and validation losses with respect to iterations, and Figure \ref{acc_iter} shows the training and validation accuracies with respect to iterations. It can be seen that after $60,000$ iterations the training loss and training accuracy start saturating indicating the convergence the network. The validation accuracy also reaches more than $99\%$ after $60,000$ iterations and saturates. We stop the training process at $70,000$ iterations and save the final parameters of the model for the future use. To test the performance of the model, the trained model was tested on the test set which consists of $50,000$ IP pairs and $50,000$ DP pairs. It is to be noted that, the test image patches also come from the same four classes only. On this test set, the model achieves an accuracy of $99.38\%$. This experiment shows that the proposed siamese network can discriminate the different types of image editing operations with a very high accuracy.

Another experiment is carried out to compare our method with Bayar and Stamm's method. In this experiment, we check the ability of the proposed method in classifying each of the four manipulation types individually. For this, we have created four different test sets as follows: for the Gaussian blurring test set, $50,000$ IP pairs are created by taking both images of the pairs from the Gaussian blurring only, and $50,000$ DP pairs are created by taking one image from the Gaussian manipulation and the other from any of the rest three classes. The test sets for the original, the median filtering, and the resampling cases are also created following the similar procedure. We have checked the classification accuracies of the already trained siamese network on these four test sets. For comparison, we have implemented Bayar and Stamm's method and tested its performance on these test images. It should be noted that the size of image patches used in this experiment is $150\times150$ as opposed to $227\times227$ used in the paper \cite{Bayar_CNN}. The classification results are shown in Table \ref{acc_diff_class}. The proposed method classifies the original, the Gaussian blurred, the median filtered and the resampled patches with accuracies of $99.35\%$, $99.51\%$, $99.64\%$ and $99.26\%$ respectively, whereas Bayar and Stamm's method classified with accuracies of $98.70\%$, $99.80\%$, $98.85\%$ and $99.13\%$ respectively. These results show that except the Gaussian manipulation, the proposed siamese network outperforms the CNN method \cite{Bayar_CNN} for all other manipulations.

\begin{table}[]
	\centering
	\caption{Classification accuracies on different manipulation classes}
	\label{acc_diff_class}
	\begin{tabular}{|c|c|c|}
		\hline
		\textbf{Manipulation} & \textbf{Proposed Method} & \textbf{Bayar and Stamm \cite{Bayar_CNN}} \\ \hline \hline
		Original     & \textbf{99.35}            & 98.70           \\ \hline
		Gaussian blurring    & 99.51           & \textbf{99.80}            \\ \hline
		Median filtering     & \textbf{99.64}           & 98.75           \\ \hline
		Resampling   & \textbf{99.26}            & 99.13           \\ \hline
	\end{tabular}
\end{table}

\begin{table}[]
	\centering
	\caption{Classification accuracies on manipulations not present in the training stage}
	\label{gen_acc}
	\begin{tabular}{|c|c|}
		\hline
		\textbf{Manipulation}                          & \textbf{Accuracy (\%)} \\ \hline \hline
		AWGN ( $\sigma = 2$)       & 96.61          \\ \hline
		Gamma correction ( $\gamma= 1.5$) & 95.24          \\ \hline
	\end{tabular}
\end{table}

The next experiment is carried out to see the ability of the proposed method in discriminating manipulations not present at the training phase. We have trained the network using the image patch pairs coming from the four classes as already mentioned. We test it on a set which contains images coming from a different type of manipulation. For this, we considered two different manipulation classes obtained by corrupting the images with AWGN and applying gamma correction operations on the image patches. For the AWGN case, we created $50,000$ IP pairs by taking both images of a pair from AWGN class only, and $50,000$ DP pairs are created by taking one image of the pair from AWGN and the other comes from one of the four classes, i.e. the original, the Gaussian blurring, the median filtering, and the resampling. On this set, the network achieved a classification accuracy of $96.61\%$. Similarly, to see the generalization ability of the network on gamma correction class, we created $50,000$ IP and $50,000$ DP test pairs in the same manner, and tested the network on it. In this case, the network achieved an accuracy of $95.24\%$. From these results, it is evident that the network can discriminate images coming from different types of manipulations, even if the images come from manipulation classes not used in the training stage. Table \ref{gen_acc} summarises the generalisation accuracy of the method. It should be noted that the existing universal forensics methods \cite{Qiu_universal}, \cite{fan_WIFS}, \cite{Bayar_CNN} cannot be applied in this case. This is because they can only classify images to one of the manipulations present in the training stage. This is a huge advantage of the proposed method over the state-of-the-art.

\section{Conclusions and Future Work}

In this paper, we proposed a novel image forensics method which can discriminate different types of image manipulations carried out on images. The proposed method employs a deep siamese CNN which takes a pair of image patches as input and decides whether they are identically or differently processed. That is, instead of classifying image patches to some fixed classes (types) of manipulations, the proposed method checks whether two image patches are processed through the same operation or not. Because of this, the proposed method can even discriminate image manipulations which were not present in the training stage. The experimental results show that the proposed method can differentiate between different types of manipulations with good accuracies.

The future work will involve further exploration of the universal nature of the proposed method. Also, checking the effectiveness of the proposed method in detecting real splicing forgeries is another task included in the future work.

\bibliographystyle{IEEEtran}
\bibliography{reference}

%

%
%
%




\end{document}